\documentclass[11pt]{article}
\usepackage{wustyle}
\usepackage[]{geometry}
\usepackage{fullpage}

\title{Embedding Inequalities for Barron-type Spaces}
\date{\vspace{-1em}}

\author{Lei Wu~\footnote{
School of Mathematical Sciences, Peking University. Email: \texttt{leiwu@math.pku.edu.cn}.}
}
\date{\today}

\begin{document}

\maketitle 
\begin{abstract}
An important problem in machine learning theory is to understand the approximation and generalization properties of two-layer neural networks in high dimensions.   To this end, researchers have introduced the Barron space $\mathcal{B}_s(\Omega)$ and the spectral Barron space $\mathcal{F}_s(\Omega)$, where the index $s\in [0,\infty)$ indicates the smoothness of functions within these spaces and $\Omega\subset\mathbb{R}^d$ denotes the input domain. However, the precise relationship between the two types of Barron spaces remains unclear. In this paper, we establish a continuous embedding between them as implied by the following inequality:
for any $\delta\in (0,1), s\in \NN^{+}$ and $f: \Omega \mapsto\mathbb{R}$, it holds that  
\[
\delta \|f\|_{\mathcal{F}_{s-\delta}(\Omega)}\lesssim_s \|f\|_{\mathcal{B}_s(\Omega)}\lesssim_s \|f\|_{\mathcal{F}_{s+1}(\Omega)}.
\]
 Importantly, the constants do not depend on the input dimension $d$, suggesting that the embedding is effective in high dimensions. Moreover, we also show that the lower and upper bound are both  tight.
\end{abstract}

\section{Introduction}
A (scaled) two-layer neural network is given by
\begin{align}\label{eqn: 2lnn}
f_m(x;\theta) = \frac 1 m\sum_{j=1}^m a_j \sigma(w_j^Tx + b_j),
\end{align}
where $\sigma: \RR\mapsto\RR$ is a nonlinear activation function; $a_j,b_j\in\RR, w_j\in\RR^d$, $\theta=\{(a_j,w_j,b_j)\}_{j=1}^m$; $m$ and $d$ denote the network width and  the input dimension, respectively. 
The extra scale factor in \eqref{eqn: 2lnn} is introduced to facilitate our subsequent analysis and it does not change the network's approximation power. Additionally, throughout this paper, we assume the input domain $\Omega\subset\RR^d$  to be compact  and focus on the case of activation function ReLU$^s$ with $s\geq 0$: 
\begin{equation*}
\sigma(z)=\max(0,z)^s.
\end{equation*}
The cases of $s=0$ and $s=1$ correspond to the Heaviside step function and vanilla ReLU function, respectively. The case of $s\geq 2$ has also found applications in solving PDEs \cite{weinan2018deep,xu2020finite,li2019better} and natural language processing \cite{so2021searching}.

\cite{cybenko1989approximation} showed that functions in $C(\Omega)$ can be approximated arbitrarily well by two-layer neural networks  with respect to the uniform metric. However,  the approximation can be arbitrarily slow.  
 \cite{pinkus1999approximation} expanded on this by showing that for  functions belonging in $C^k(\Omega)$, the approximation by two-layer neural networks can achieve a rate of $O(m^{-k/d})$. This rate, unfortunately,  is subject to the curse of dimensionality since it diminishes as $d$ increases. These suggest that mere continuity and smoothness are not sufficient to ensure an efficient approximation in high dimensions. 
 Then it is natural to ask: what kind of regularity can ensure the efficient approximation by two-layer neural networks? 
Before proceeding to review previous studies attempting to answer this question. We  need a dual norm for handling the compactness of input domain. 

\begin{definition}[\cite{barron1993universal}]
Given a compact set $\Omega$, we define 
$
    \|v\|_{\Omega} = \sup_{x\in \Omega} |v^Tx|.
$
\end{definition}

We begin by considering the spectral Barron spaces \cite{siegel2020approximation,xu2020finite,siegel2023characterization,caragea2020neural}, which are  defined as follows:
\begin{definition}\label{def: spectral-barron}
Let $\cX\subset\RR^d$ be a compact domain. For $f: \cX\mapsto\RR$ and $s\geq 0$, define
\begin{equation*}
    \|f\|_{\cF_s(\cX)} = \inf_{f_e|_{\cX}=f} \int_{\RR^d} (1+\|\xi\|_{\cX})^s |\hat{f_e}(\xi)|\dd \xi,
\end{equation*}
where the infimum is taken over all  extensions of $f$. Let 
$$
\cF_s(\cX):=\{f:\cX\mapsto\RR \, :\, \|f\|_{\cF_s(\cX)}<\infty\}.
$$
Then, the spectral Barron space is defined as $\cF_s(\Omega)$ equipped with the $\|\cdot\|_{\cF_s(\cX)}$ norm.
\end{definition}
In the above definition, we consider measure-valued Fourier transform as done in \cite{barron1993universal}. It is worth noting that Definition \ref{def: spectral-barron} bears resemblance to the Fourier-based characterization of Sobolev spaces, denoted as $\|f\|_{H_s}^2=\int_{\RR^d} (1+\|\xi\|)^s |\hat{f}(\xi)|^2\dd \xi$. The major distinction lies in the fact that the moment in Definition \ref{def: spectral-barron} is calculated with respect to $|\hat{f}(\xi)|$ instead of $|\hat{f}(\xi)|^2$.

It was proved in \cite{xu2020finite} that if $\|f\|_{\cF_{s}(\Omega)}<\infty$, 
then functions in $\cF_{s}(\Omega)$ can be approximated by two-layer ReLU$^{s-1}$ networks without suffering the curse of dimensionality.  Specifically, the approximation error obeys the  Monte-Carlo error rate $O(m^{-1/2})$, where $m$ denotes the network width.
The special case of $s=1$ was first  considered in the pioneer work of  Andrew Barron \cite{barron1993universal}. Subsequently, the case of $s=2$ was studied in \cite{breiman1993hinging,klusowski2016risk}. More recently, the extension to general positive integer $s$ was  provided in 
\cite{siegel2020approximation,xu2020finite,caragea2020neural}.


The Fourier-based characterization, while explicit, is not necessarily tight as it may exclude functions that can be effectively approximated by two-layer neural networks.  \cite{ongie2019function,parhi2021banach} considered similar characterizations based on Radon transform instead of  Fourier transform, which can yield a tight characterization for the case of $d=1$. Moreover, \cite{ma2019priori, weinan2021barron} offered a probabilistic generalization of Barron's analysis \cite{barron1993universal}. In these studies, functions satisfying the following expectation representation are taken into consideration:
\begin{equation}\label{eqn: expectation}
    f_\rho(x) = \EE_{(a,w,b)\sim\rho} [a\sigma(w^Tx+b)], \quad \forall x\in \cX,
\end{equation}
where $\rho\in \cP(\RR\times\RR^d\times \RR)$.
This can be obtained from \eqref{eqn: 2lnn} by taking $m\to\infty$ and applying the law of large numbers.  One can view $f_\rho$ as an infinitely-wide two-layer neural network. It is important to note that the expectation representation in \eqref{eqn: expectation} only needs to hold in  $\cX$ instead of  the entire space $\RR^d$. Accordingly, the (probabilistic) Barron spaces are defined as follows:

\begin{definition}\label{def: Barron}
 Given $s\geq 0$ and $f:\cX\mapsto\RR$, let $A_f:=\{\rho\in \cP(\RR\times\RR^d\times\RR)\,:\, f_{\rho}|_{\cX}=f\}$. Then, the Barron norm of  $f$ and the associated Barron space is defined by 
\[
    \|f\|_{\cB_s(\cX)}:=\inf_{\rho \in A_f} \EE_{(a,w,b)\sim\rho}[|a|(\|w\|_{\cX}+|b|)^s].
\] 
Let $$
\cB_s(\cX) =\{ f:\cX\mapsto\RR \,:\, \|f\|_{\cB_s(\Omega)}<\infty\}.
$$
Then the Barron space is defined as $\cB_s(\cX)$ equipped with the $\|\cdot\|_{\cB_s(\Omega)}$ norm.
\end{definition}
The above definition is a slight generalization of  the one originally proposed in \cite{weinan2021barron}, where only the case of $s=1$ is considered. Following the proofs in \cite{weinan2021barron} and \cite{ma2019priori}, one can easily show that approximating and estimation error for learning functions in $\cB_s$  with two-layer $\relu^s$ networks follow the Monte-Carlo rates $O(m^{-1/2})$ and $O(n^{-1/2})$, respectively. Here $n$ denotes the number of training samples.  Recently,  \cite{siegel2022sharp} established a sharper approximation rate  of $O(m^{-1/2-(s+1/2)/d})$. However, it is important to note that this rate improvement is less significant in high dimensions  and additionally, the hidden constants in \cite{siegel2022sharp} may have an exponential dependence  on $d$.
Compared with the Fourier-based characterization in Definition \ref{def: spectral-barron}, the above expectation-based characterization is  more natural and complete. Specifically, \cite{weinan2021barron} provided an inverse approximation theorem, showing that if $f$ can be approximated by two-layer $\relu$ networks with bounded path norm \cite{neyshabur2015norm}, it must lie in $\cB_1(\Omega)$. 

\subsection{Our Contribution}
Recently,  Barron-type spaces defined above have been adopted to  explore various high-dimensional problems. For instance, \cite{lu2021priori-b,lu2021priori,chen2021representation,weinan2022some} established some regularity theories of high-dimensional PDEs with Barron-type spaces. Hence, it is natural to ask: what is the relationship between them?  \cite{barron1993universal,breiman1993hinging,klusowski2016risk,xu2020finite} already showed that  $\cF_{s+1}(\Omega)\subseteq \cB_s(\Omega)$. Moreover, \cite{weinan2022representation} provided a specific example showing that $\cF_2(\Omega)\subsetneq \cB_1(\Omega)$, implying that $\cB_1(\Omega)$ is strictly larger than $\cF_2(\Omega)$. Along this line of work, our major contribution is the following precise embedding result:

\begin{theorem}\label{thm: fourier-lowerbound}
Let $\cX\subset \RR^d$ be a compact set.
For  any $s\in \NN^{+}, f\in \cB_s(\Omega), \delta\in (0,1)$, we have 
\begin{equation*}\label{eqn: spectra-norm-barron}
     \delta \|f\|_{\cF_{s-\delta}(\Omega)} \lesssim_s   \|f\|_{\cB_s(\Omega)}  \lesssim_s  \|f\|_{\cF_{s+1}(\Omega)},
\end{equation*}
where $s$ in the upper bound can take the value of $0$.
\end{theorem}
Note that the hidden embedding constants  depend solely on the value of $s$. This suggests that the embedding revealed in Theorem \ref{thm: fourier-lowerbound}  is effective in high dimensions.  Additionally,     
as per our current proof, the smoothness index $s$ is required to be a  positive integer, though $\cB_{s}(\Omega)$ is defined for any  $s$ in the range of  $[0,\infty)$. However,   we conjecture that analogous results  would apply for any $s \in (0,+\infty)$ as discussed in Remark \ref{remark: 2}, which we leave for future work. 




Additionally, we would like to clarify that the upper bound in Theorem \ref{thm: fourier-lowerbound}  has already been implicitly established in previous works. Specifically, the case of $s=0$ was proven  in the pioneering work of  Andrew Barron \cite{barron1993universal} albeit presented in a different form. Subsequently, the analysis was extended to the case of $s=1$ in \cite{breiman1993hinging, klusowski2016risk}, and further generalized to arbitrary non-negative integer values of $s$ in \cite{xu2020finite, siegel2020approximation}.  Our major contribution is the lower bound, which is critical for establishing the embedding between the two types of Barron spaces and the proof is presented in Section \ref{sec: proof-thm1}. In Theorem \ref{thm: fourier-lowerbound}, the upper bound is stated for the sake of completeness.

We mention that \cite{meng2022new} establishes the embedding among spectral Barron spaces and some classical spaces such as the Sobolev space,  Besov space, and Bessel potential space. In contrast, we focus on the embedding between the Barron spaces and spectral Barron spaces.

\begin{figure}[!h]
\centering
\includegraphics[width=0.35\textwidth]{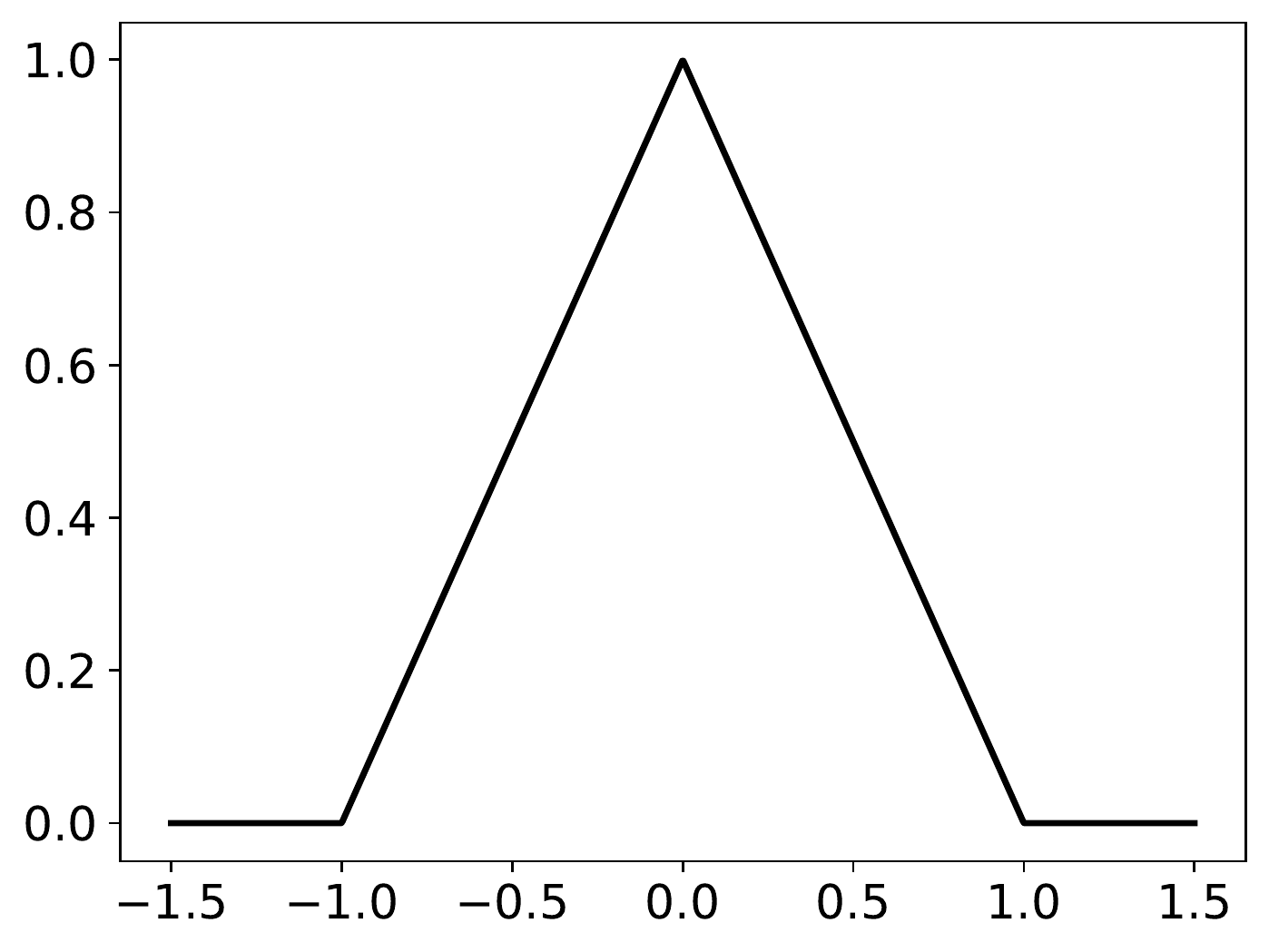}
\vspace*{-3mm}
\caption{\small The triangular function $t(x):=\max(1-|x|,0)$.}
\label{fig: triangluar-funct}
\end{figure}

\paragraph*{Tightness.}
For the upper bound, \cite[Proposition 7.4]{caragea2020neural} shows that when $\Omega$ has nonempty interior, if $\cF_{s}(\Omega)\subset \cB_{1}(\Omega)$, then we must have $s\geq 2$. This implies that the upper bound is tight.  
The following proposition shows that the lower bound in Theorem \ref{thm: fourier-lowerbound} is also tight in the sense that the value of $\delta$ cannot be taken to zero.
\begin{proposition}\label{pro: tightness}
Let $\cX=[-1,1]$ and $f(x)=\max(1-|x|,0)$ for $x\in \cX$. Then,
\begin{align*}
     \|f\|_{\cB_1(\Omega)} \leq 3, \qquad \|f\|_{\cF_1(\Omega)} = +\infty.
\end{align*}
\end{proposition}
Let $t(x) := \max(1-|x|,0)$ for any $x\in\RR$ be the triangular hat function (see Figure \ref{fig: triangluar-funct}). We have 
$
    \hat{t}(\xi) = \frac{1-\cos(\xi)}{\pi\xi^2}.
$
Note that $t(\cdot)$ is a zero extension of $f$ and $\int_{\RR} (1+|\xi|)|\hat{t}(\xi)|\dd \xi=+\infty$. However, this does not directly imply $\|f\|_{\cF_1(\Omega)}=\infty$, since the spectral Barron norm is defined by taking the infimum over all  possible extensions. We refer to Section \ref{sec: proof-lower-bound} for a rigorous proof.

\section{Proofs}
\label{sec: proof}
\paragraph*{Notation.} We use $X\lesssim_{\alpha} Y $ to denote  $X\leq C_\alpha Y$ where $C_\alpha$ is a positive constant that depends only on $\alpha$.
For a vector $v$, let $\|v\|_p=(\sum_j v_j^p)^{1/p}$.
Let $\SS^{d-1}=\{x\in\RR^d:\|x\|_2=1\}$ and 
$
\SS^{d-1}_{\Omega}=\{x\in\RR^d\,:\, \|x\|_{\Omega}=1\}.
$
Denote by  $\indicator_{S}$ the indicator function of the set $S$, satisfying $\indicator_{S}(x)=1$ for $x\in S$, and $0$ otherwise. For a metric space $X$, denote by $\cP(X)$ the set of probability measures over $X$.

    Throughout this paper, we define Fourier transform as follows 
\[
    \hat{f}(\xi) = \frac{1}{(2\pi)^d} \int_{\RR^d}  e^{-i \xi^Tx} f(x)\dd x,
\]
and the inverse Fourier transform is given by 
\[
    f(x) = \int_{\RR^d}  e^{i\xi^Tx} \hat{f}(\xi)\dd \xi.
\]
 Note that in these definitions, the terms $f(x)\dd x$ and $\hat{f}(\xi)\dd \xi$ should be interpreted as a finite measure in a broad sense. Moreover, we will use the identity: for $d=1$,
$$
    \delta(\xi) = \frac{1}{2\pi}\int_\RR e^{-i\xi x}\dd x.
$$

Before proceeding to the proof, we first clarify several important issues that might be ignored. Both types of Barron functions are defined on a compact domain $\cX$ instead of the whole space $\RR^d$ and thus, Barron norms depend on the underlying domain $\cX$. When estimating Barron norms, one need to be careful with the choice of extensions.
A naive extension may yield a significantly loose bound of the $\cF_s(\Omega)$ norm \cite{domingo-enrich2022tighter} and $\cB_s(\Omega)$ norm.


\subsection{Proof of Theorem \ref{thm: fourier-lowerbound}}
\label{sec: proof-thm1}

We start by  considering the case of single neurons.
For any $w\in\SS^{d-1}_\Omega,b\in\RR$, the single neuron  $\sigma_{w,b}: \Omega\mapsto\RR$ is given by $\sigma_{w,b}(x)=\sigma(w^Tx+b)$. Note that the domain of $\sigma_{w,b}$ is $\cX$ instead of $\RR^d$. In particular,  when $d=1$ and $w=1$, we write $\sigma_{b}=\sigma_{w,b}$ for simplicity. 
The following lemma  characterizes the Fourier transform of a single neuron.

\begin{lemma}\label{lemma: single-neuron-ext}
Let  $\sigma_{w,b}: \cX \mapsto \RR$ with $\|w\|_\Omega =1$ be a single neuron and $g:\RR\mapsto\RR$ be any extension of $1_{[-1,1]}$.  Then, $G_{w,b}(x):= \sigma_{w,b}(x)g(w^Tx)$ is an extension of $\sigma_{w,b}$, satisfying
\begin{equation}\label{eqn: d-1-single}
    \int_{\RR^d} (1+\|\xi\|_{\Omega})^s |\widehat{G}_{w,b}(\xi)|\dd\xi =  \int_{\RR} (1+|v|)^s |\hat{h}_{\sigma,b}(v)|\dd v,
\end{equation}
where $h_{\sigma,b}(z)=\sigma(z+b)g(z)$ is an extension of $\sigma_{b}: [-1,1]\mapsto\RR$.
\end{lemma}

\begin{proof}
Let $Q=(w,w_2,\dots,w_d)^T\in\RR^{d\times d}$ with $w_2,\dots,w_d$ being orthonormal and $w_i^Tw=0$ for $i=2,\dots,d$. Then, by letting $\bar{\xi}=(Q^{-1})^T\xi$, we have
\begin{align}
\notag \widehat{G}_{w,b}(\xi) &= \frac{1}{(2\pi)^d}\int_{\RR^d} \sigma(w^Tx+b) g(w^Tx)e^{-i \xi^Tx} \dd x\\
\notag &=\frac{1}{(2\pi)^d}\int_{\RR^d} \sigma(y_1+b)g(y_1)e^{-i \xi^T Q^{-1}y}\frac{1}{|\det Q|} \dd y \qquad  (y=Qx) \\
\notag &=\frac{1}{|\det Q|}\left(\frac{1}{2\pi}\int_{\RR^d} \sigma(y_1+b)g(y_1) e^{-i \bar{\xi}_1 y_1}\dd y_1 \right)\prod_{j=2}^d \delta(\bar{\xi}_j) \\ 
&= \frac{1}{|\det Q|}\hh_{\sigma,b}(\bar{\xi_1}) \prod_{j=2}^d \delta(\bar{\xi}_j).
\end{align}

Now, we have
\begin{align}
 \notag   \int_{\RR^d}(1+\|\xi\|_\Omega)^s |\widehat{G}_{w,b}(\xi)|\dd \xi &= \int_{\RR^d}(1+\|Q^T\bar{\xi}\|_\Omega)^s \frac{1}{|\det Q|}|\hh_{\sigma,b}(\bar{\xi}_1)| \prod_{j=2}^d \delta(\bar{\xi}_j) |\mathrm{det}Q|\dd \bar{\xi} \\
\notag &=\int_{\RR^d}(1+\|\bar{\xi}_1w + \sum_{j=2}^d \bar{\xi}_j w_j\|_\Omega)^s |\hh_{\sigma,b}(\bar{\xi}_1)| \prod_{j=2}^d \delta(\bar{\xi}_j) \dd \bar{\xi} \\
\notag    &=\int_{\RR}(1+\| w\|_\Omega |\bar{\xi}_1|)^s |\hh_{\sigma,b}(\bar\xi_1)| \dd \bar\xi_1\\ 
&=\int_{\RR}(1+ |\bar{\xi}_1|)^s |\hh_{\sigma,b}(\bar\xi_1)| \dd \bar\xi_1
\end{align}
where the first step use $\xi=Q^T\bar{\xi}$ and the last step is due to $\|w\|_\Omega=1$.
\end{proof}


The above lemma provides a way to estimating spectral Barron norms of single neurons. 
What remains is to determine an extension $g$ such that the right hand side of Eq.~\eqref{eqn: d-1-single} to be as small as possible. To this end, we first consider the one-dimensional case.

When $d=1$, for any $b\in\RR$, let $\sigma_b=\sigma(\cdot+b): [-1,1]\mapsto\RR$. When it is clear from the context, we also use $\sigma_b$ denote the single neuron define on the entire space.
Let $\chi:\RR\mapsto\RR$ be a smooth cutoff function, satisfying $\chi\in C_c^{\infty}(\RR)$ and  $\chi(z)=1$ for any $z\in [-1,1]$ and $\supp \chi=[-2,2]$.   Given any $b\in\RR$, we shall consider the following extension of a single neuron:
\[
h_{\sigma,b}(z)=\chi(z) \sigma_b(z):\RR\mapsto\RR.
\] 

\begin{lemma}\label{lemma: 1d-extension}
Let $\sigma(z)=\max(0,z)^s$ with $s\in \NN$. Then,
$ 
    |\hat{h}_{\sigma,b}(\xi)|\lesssim_s (1+|b|)^s/(1+|\xi|)^{s+1}.
$
\end{lemma}

\begin{remark}\label{remark: 2}
The proof uses explicitly the condition of $s\in \NN$. However, according to the relationship between the smoothness of a function and the decay of the Fourier transform, we anticipate that the same result holds for any $s\in [0,\infty)$.
\end{remark}

\begin{proof}
Using the product rule, we have for any $k\in \NN^{+}$
\[
    h^{(k)}_{\sigma,b}(z) = \sum_{i=0}^{k} \binom{k}{i}\sigma_b^{(i)}(z) \chi^{(k-i)}(z)
\]
and  prove the theorem for the following  two cases separately. Without lose of generality, we consider here only the case of $b\geq 0$. When $b$ is negative, the proof is similar. 


\underline{\bf The case  of $b\geq 2$:~} In this case, $h_{\sigma,b}(\cdot)=\sigma_b(\cdot)\chi(\cdot)\in C^{\infty}(\RR)$. Without loss of generality, we consider the case of $b\geq 1$, for which
\begin{equation*}
h_{\sigma,b}(z)=\begin{cases}
0 & \text{ if } z< -2\\ 
(z+b)^s\chi(z) & \text{ if } z\in [-2,2]\\
0 & \text{ if } z>2
\end{cases}
\end{equation*}
\begin{itemize}
\item When $z\in [-2,2]$,  we have $\sigma_b^{(k)}(z)=0$ for $k>s$ and $|\sigma^{(k)}_b(z)|\lesssim_s (1+|b|)^s$ for $k\leq s$. Hence, for any $k\in \NN$, we have for any $k\in \NN$
\begin{align*}
    |h^{(k)}_b(z)|&=\sum_{i=0}^{\min\{k,s\}} \binom{k}{i}\sigma_b^{(i)}(z) \chi^{(k-i)}(z)\\
    &\lesssim_s \sum_{i=0}^{\min\{k,s\}} |\sigma_b^{(i)}(z)| |\chi^{(k-i)}(z)|\lesssim_s \sum_{k=0}^{\min\{k,s\}} |\sigma_b^{(k)}(z)| \lesssim_s (1+|b|)^s.
\end{align*}
\item When $|z|>2$, $|h^{(k)}_{\sigma,b}(z)|=0$ for any $k\in \NN$.
\end{itemize}
Combining two cases leads to  for any $k\in \NN$, 
\[
\|h^{(k)}_{\sigma,b}\|_{L^1(\RR)}=\int_{-2}^2 |h^{(k)}_{\sigma,b}(z)|\dz \lesssim_s (1+|b|)^s.
\]
This implies
\begin{equation}\label{eqn: decay-1}
    |\hh_b(\xi)|=\left|\frac{1}{2\pi (-i\xi)^{s+1}}\int_{\RR} h_{\sigma,b}^{(s+1)}(x)e^{-i\xi x}\dx\right|\lesssim_s \frac{(1+|b|)^s}{|\xi|^{s+1}}.
\end{equation}

\underline{\bf The case of $0\leq b<2$:~}
In this case, $h_{\sigma,b}$ is piecewise smooth, given by 
\[
    h_{\sigma,b}(z)=\begin{cases}
    0 & \text{ if }  z\leq -b \\ 
    (z+b)^s\chi(z) & \text{ if } z>-b.
    \end{cases}
\]
Consequently, $h_{\sigma,b}^{(s)}(\cdot)$ is bounded  and has only one discontinuity point at $z=-b$. By adopting the product rule in a way similar as the above, it is not hard to show that for all $k\in \NN$,
\begin{equation}\label{eqn: x1}
\begin{aligned}
    h^{(k)}_{\sigma,b}(z)&=0 \text{ for all } z\in (-\infty, -b)\cup [2,+\infty)\\ 
    |h^{(k)}_{\sigma,b}(z)|&\lesssim_s 1 \text{ for all } z\in (-b,2].
\end{aligned}
\end{equation}
and $\lim_{z\to (-b)^{+}}h^{(s)}_{\sigma,b}(z)$ exists with 
$ 
|\lim_{z\to (-b)^{+}}h^{(s)}_{\sigma,b}(z)| \lesssim_s 1.
$

Noting that
\begin{align*}
\hh_{\sigma,b}(\xi) &= \frac{1}{2\pi}\int_{-\infty}^\infty h_{\sigma,b}(z) e^{-i \xi z}\dz=\frac{1}{2\pi (-i\xi)^s}\int_{-\infty}^\infty h_{\sigma,b}^{(s)}(z) e^{-i \xi z}\dz \\
&= \frac{1}{2\pi (-i\xi)^s}\int_{-b}^2 h_{\sigma,b}^{(s)}(z) e^{-i \xi z}\dz \\
&=  \frac{1}{2\pi (-i\xi)^s}\left(\frac{e^{-i\xi z}}{-i\xi}h_{\sigma,b}^{(s)}(z)\Big |_{-b}^2 + \int_{-b}^2 h_{\sigma,b}^{(s+1)}(z)\frac{e^{-i\xi z}}{i\xi} \dz \right)
\end{align*}
and applying \eqref{eqn: x1}, we have
\begin{align}\label{eqn: xx2}
|\hh_{\sigma,b}(\xi)|\lesssim_s \frac{1}{|\xi|^{s+1}} \left( 1+ \int_{-b}^2\dz\right)\lesssim  \frac{1}{|\xi|^{s+1}}\lesssim_s \frac{(1+|b|)^s}{|\xi|^{s+1}},
\end{align}
where the last step uses the assumption of $|b|\leq 2$.

On the other hand, when $|\xi|\leq 1$, we have for any $b\in\RR$ that
\begin{equation}\label{eqn: xx3}
|\hh_{\sigma,b}(\xi)|\leq \frac{1}{2\pi}\int_{\RR} |h_{\sigma,b}(z)e^{-i\xi z}\dz|\leq \frac{1}{2\pi}\int_{-2}^2 |h_{\sigma,b}(z)|\dz\lesssim_s (1+|b|)^s.
\end{equation}
Then, combining \eqref{eqn: xx3} with \eqref{eqn: decay-1} and \eqref{eqn: xx2}  yields
$
    |\hh_{\sigma,b}(\xi)|\lesssim_s (1+|b|)^s/(1+|\xi|)^{s+1}.
$
\end{proof}

\begin{lemma}\label{lemma: single-neuron-moment}
Given any $w\in \SS^{d-1}_\Omega, b\in\RR$, consider the extension $H_{w,b}(x):=\sigma(w^Tx+b)\chi(w^Tx)$. Then, for any $\delta\in (0,1)$, we have 
\begin{equation}\label{eqn: extension}
    \int_{\RR} (1+\|\xi\|_\Omega)^{s-\delta} |\hat{H}_{w,b}(\xi)|\dd \xi \lesssim_s \delta^{-1}(1+|b|)^s.
\end{equation}
\end{lemma}

\begin{proof}
By Lemma \ref{lemma: single-neuron-ext} and Lemma \ref{lemma: 1d-extension}, we have 
\begin{align*}
    \int_{\RR} (1+\|\xi\|_\Omega)^{s-\delta} |\hat{H}_{w,b}(\xi)|&=\int_{\RR} (1+|v|)^{s-\delta}|\hh_{\sigma,b}(v)|\dd v\\
    &\lesssim_s\int_{\RR} (1+|v|)^{s-\delta} \frac{(1+|b|)^s}{(1+|v|)^{s+1}}\dd v\\
    &\lesssim_s (1+|b|)^s \int_{\RR}\frac{1}{(1+|v|)^{1+\delta}}\dd v\lesssim_s \frac{(1+|b|)^s}{\delta}.
\end{align*}
\end{proof}

\paragraph*{The proof of Theorem \ref{thm: fourier-lowerbound}.} We are now ready to prove the main theorem.
For any $f\in \cB_s(\Omega)$ and $\varepsilon>0$, there exists $\rho_\epsilon\in \cP(\RR\times \SS_{\Omega}^{d-1}\times\RR)$ such that 
\begin{align*}
    f(x) &= \int a\sigma(w^Tx+b)\dd\rho_\epsilon(a,w,b),\quad \forall x\in \cX\\
    \int |a|(1+|b|)^s \dd \rho_\epsilon(a,w,b)&\leq \|f\|_{\cB_s(\Omega)}+\epsilon,
\end{align*}
where we have used the positive homogeneity of ReLU$^s$ and set $w\in \SS_{\Omega}^{d-1}$.
Let 
\[
    f_e(x) = \int a\sigma(w^Tx+b)\chi(w^Tx)\dd\rho_\epsilon(a,w,b)=\int a H_{w,b}(x)\dd\rho_\epsilon(a,w,b),\quad \forall x\in\RR^d,
\]
where $H_{w,b}:\RR^d\mapsto \RR$ is the extension of $\sigma_{w,b}$ defined in Lemma \ref{lemma: single-neuron-moment}.
Then, $f_e$ is an extension of $f$ and satisfies
\[
    \hf_e(\xi) = \int a \hat{H}_{w,b}(\xi)\dd\rho_\epsilon(a,w,b).
\]

According to Lemma \ref{lemma: single-neuron-moment}, we have
\begin{align*}\label{eqn: xxxx}
\notag \int_{\RR^d} (1+\|\xi\|_{\Omega})^{s-\delta} |\hf_e(\xi)|\dd\xi &\leq \int |a|\left(\int_{\RR^d} (1+\|\xi\|_{\Omega})^{s-\delta} |\hat{H}_{w,b}(\xi)|\dd\xi\right)\dd\rho_\epsilon(a,w,b)\\
&\lesssim_s \int |a| \frac{(1+|b|)^s}{\delta}\dd \rho_\epsilon(a,w,b)\leq \frac{1}{\delta} (\|f\|_{\cB_s(\Omega)}+\epsilon).
\end{align*}
By the definition of spectral Barron norm,  it follows that
$
    \|f\|_{\cF_{s-\delta}(\Omega)}\lesssim_s \delta^{-1}(\|f\|_{\cB_s(\Omega)} + \epsilon).
$
Taking $\epsilon\to 0$ completes the proof.

The converse direction follows from \cite{klusowski2016risk,xu2020finite,siegel2023characterization}.
\qed

\subsection{Proof of Proposition \ref{pro: tightness}}
\label{sec: proof-lower-bound}

\begin{proof}
Notice that $f(\cdot)$ can be exactly represented as a two-layer neural network for $x\in [-1,1]$:
\[
    f(x) = \sigma(1)-\sigma(x)-\sigma(-x). 
\]
Hence, $f$ is a Barron function and obviously, $\|f\|_{\cB_1(\Omega)}\leq 3$.

What remains is to show that  $\int (1+|\xi|)|\hf_e(\xi)|\dd\xi=\infty$ holds for any  extension $f_e$. Suppose, to the contrary, that there exists an extension $f_e$ such that   $\int (1+|\xi|) |\widehat{f}_e(\xi)|\dd \xi<\infty$. Then $\hf_e\dd \xi$ represents a finite measure over $\RR^d$ and $f_e$ is continuous in $\cX$.  By the  Fourier inverse theorem, we have 
\[
    f(x) = \int  e^{i \xi x} \hf_e(\xi) \dd x \quad\forall\, x\in \cX.
\]
For any $x\in (-1/2,0)\cup(0,1/2)$ and sufficiently small $\delta$, 
\begin{align}\label{eqn: x}
\frac{f(x+\delta)-f(x)}{\delta} = \int  e^{i \xi x} \frac{e^{i\xi \delta}-1}{\delta} \hf_e(\xi) \dd \xi.
\end{align}
The integrand on the right side of \eqref{eqn: x} is bounded by $|\xi||\hat{f}_e(\xi)|$, which is  integrable by the assumption.  Consequently, by the dominated convergence theorem, for $x\in (-1/2,0) \cup (0,1/2)$, we have
\[
    f'(x) =\lim_{\delta\to 0} \frac{f(x+\delta)-f(x)}{\delta}= \int  e^{i \xi x} \lim_{\delta\to 0}\frac{e^{i\xi \delta}-1}{\delta} \hf_e(\xi)\dd \xi= \int  i\xi e^{i \xi x} \hf_e(\xi) \dd \xi.
\]
Again, by dominated convergence theorem and taking $x\to 0$, 
\[
\lim_{x\to 0} f'(x) =\lim_{x\to 0}   \int  i\xi e^{i \xi x} \hf_e(\xi) \dd \xi = \int \lim_{x\to 0} i\xi e^{i \xi x} \hf_e(\xi) \dd \xi = i \int \xi \hf_e(\xi)\dd \xi.
\]
This is contradictory to the fact that $\lim_{x\to 0}f'(x)$ does not exist. 

\end{proof}

\section{Concluding Remark}
In this paper, we establish a continuous embedding for  Barron-type spaces over compact domains. Crucially, the embedding constants do not depend on the input dimension, implying that the embedding is effective in high dimensions. We thus establish a more unifying perspective for understanding the high-dimensional approximation of two-layer neural networks. This embedding result has potential implications  for the analysis of approximating solutions of high-dimensional PDEs with two-layer neural networks \cite{chen2021representation,weinan2022some,lu2021priori-b}.

For future work, it is promising to extend our embedding result to the case of  $s\in (0,\infty)$ as discussed in Remark \ref{remark: 2}. Additionally,
our proof heavily relies on the positive homogeneity property of the ReLU$^s$ activation function. It would be interesting to extend our analysis to Barron spaces associated with general activation functions \cite{li2020complexity}. 

\paragraph*{Aknowledgement.}
We would like to thank Professor Weinan E and Dr. Jihong Long for helpful discussions and anonymous reviewers for detailed and constructive feedback.

\bibliographystyle{alpha}
\bibliography{ref}


\end{document}